\documentclass[conference]{IEEEtran}
\IEEEoverridecommandlockouts
\usepackage{cite}
\usepackage{amsmath,amssymb,amsfonts}
\usepackage{algorithmic}
\usepackage{graphicx}
\usepackage{textcomp}
\usepackage{xcolor}
\usepackage{multirow}
\usepackage{rotating}
\usepackage{hyperref}
\usepackage{xcolor}

\usepackage{soul}
\def\BibTeX{{\rm B\kern-.05em{\sc i\kern-.025em b}\kern-.08em
    T\kern-.1667em\lower.7ex\hbox{E}\kern-.125emX}}
\begin{document}
\title{ODE guided Neural Data Augmentation Techniques for Time Series Data and its Benefits on Robustness \\
{\footnotesize \textsuperscript{}}
}

\author{\IEEEauthorblockN{\textbf{Anindya Sarkar}}
\IEEEauthorblockA{\textit{} 
\textbf{IIT Hyderabad, INDIA} \\
anindyasarkar.ece@gmail.com}
\and
\IEEEauthorblockN{\textbf{Anirudh Sunder Raj}}
\IEEEauthorblockA{\textit{} 
\textbf{IIT Madras, INDIA}\\
cs17b003@cse.iitm.ac.in}
\and
\IEEEauthorblockN{\textbf{Raghu Sesha Iyengar}}
\IEEEauthorblockA{\textit{} 
\textbf{IIT Hyderabad, INDIA} \\
raghu.sesha@gmail.com}}


\IEEEoverridecommandlockouts
\IEEEpubid{\makebox[\columnwidth]{978-1-4799-7492-4/20/\$31.00~
\copyright2020
IEEE \hfill} \hspace{\columnsep}\makebox[\columnwidth]{ }} 

\maketitle

\begin{abstract}
Exploring adversarial attacks and studying their effects on machine learning algorithms has been of interest to researchers. Deep neural networks working with time series data have received lesser interest compared to their image counterparts in this context. In a recent finding, it has been revealed that current state-of-the-art deep learning time series classifiers are vulnerable to adversarial attacks. In this paper, we introduce neural data augmentation techniques and show that classifier trained with such augmented data obtains state-of-the-art classification accuracy as well as adversarial accuracy against Fast Gradient Sign Method (FGSM) and Basic Iterative Method (BIM) on various time series benchmarks.
\end{abstract}


\section{Introduction and Related Work}
Deep Neural Networks have displayed impressive results on many machine learning tasks on image (\cite{Redmon2016Yolo9k,Szegedy2016Incep,He2017MRNN,Tan2019EffNet,Zhao2019DLImageReview}), natural language processing (\cite{Sutskever2014Seq2Seq,Mikolov2013Word2Vec,Devlin2018Bert,Otter2018Survey}) and time series classification (\cite{WangTSCBaseline2017, KarimMVLSTM2019, KarimLSTMFC2019, Ismail2019TSCReview}).  However, their fragility to small adversarial perturbations is a matter of concern for researchers.  A targeted black-box attack on neural networks for image classification was formulated as an optimization problem in \cite{SzegedyTargBB2013}, and further improved in \cite{CarliniTarg2017}.  A targeted $l_{0}$ norm attack discussed in \cite{Papernotl0targ2016}, aims to minimize the number of modified pixels in an image to cause mis-classification as a particular target class.  Adversarial attacks intended to lower reliability of neural networks are also explored.  Of these, gradient based $l_{\infty}$ norm attacks such as \cite{GoodfellowFGSM2014} and \cite{KurakinBIM2016} are very popular.  Various techniques to understand and mitigate effects of adversarial perturbations have also been studied \cite{AzulaySamplingeffect2018}, \cite{RaghunathanCertDef2017},\cite{WongProvableDef2017},\cite{PapernotDisti2016},\cite{MadryMRobustGradDes2017},\cite{Sarkar2020LiCSNet}.  An excellent review of adversarial attacks on machine learning systems can be found in \cite{YevgeniyAdvMLBook2018}.  The reliability and security concerns raised by adversarial attacks have been one of the main reasons for deep neural networks not yet becoming popular with safety critical applications where the cost of failure is high.

Time series data is omnipresent and classification tasks on time series data finds its applications in health care (\cite{AbdelfattahEEGHealth2018}), power consumption monitoring (\cite{ZhengEnergy2018}), food safety (\cite{BriandetFoodSafety1996,NawrockaFoodSafety2013}), security (\cite{TobiyamaSecurity2016}) etc.  Current state-of-the-art deep neural networks can achieve impressive performance at classifying time series data (\cite{WangTSCBaseline2017,KarimMVLSTM2019,KarimLSTMFC2019}) on various datasets (\cite{ChenUCR2015,BagnallUEAMV2018}). However, these networks suffer in the same way to 
adversarial inputs as their image counterparts. A recent finding (\cite{FawazTSCAdv2019}) shows that vulnerability of state-of-the-art time series classification networks to simple adversarial attacks, such as Fast Gradient Sign Method (FGSM) and Basic Iterative Method (BIM), bring back the focus on building more robust time series classifiers. 

FGSM \cite{GoodfellowFGSM2014} uses the gradient of the loss with respect to the input data, then adjusts the input data to maximize the loss. This can be summarised using the following expression:
\begin{gather*}
    z_{adv}\> =\>z\>+\>\epsilon * sign{(\nabla_{z} J(\theta,z,y))}
\end{gather*}
where original input data and its corresponding output label are represented as $z$ and $y$ respectively, $\theta$ denotes parameters of the classifier, $J(\theta,z,y)$ denotes the loss or objective function and the obtained adversarial data is represented as $z_{adv}$. Note that $\epsilon$ is used as a multiplier to ensure small perturbations.
BIM \cite{KurakinBIM2016} extends FGSM by applying it iteratively with a small step size and clip the obtained time series elements after each step to ensure that they are in an $\epsilon$-neighborhood of the original input data.

It has been shown (\cite{FawazTSAug2018,ForestierTSAug2017}) that data augmentation increases the size and diversity of the training set resulting in improved classification accuracy on time series data.  Further, in order to achieve adversarial robustness, the classifiers should be robust to noise-corrupted data (\cite{FordNoise2019}).  In this paper, we propose data augmentation techniques for time series data, which helps in improving the robustness of the classifier against adversarial attacks.  Our contributions can be summarized as follows:
\begin{enumerate}
	\item We propose \textbf{Input gradient based Data Augmentation method}. We show that classifier trained with this augmented dataset achieves state-of-the-art classification accuracy on adversarially perturbed UCR time series datasets using FGSM and BIM attacks.
	\item We also propose \textbf{Output gradient based Data Augmentation method}. We show that classifier trained with this augmented dataset further improve the state-of-the-art classification accuracy on adversarially perturbed UCR time series datasets using FGSM and BIM attacks.
	\item Additionally, We propose a \textbf{spectral density based Data Augmentation method}. Our experimental finding suggests that, combination of Output gradient based Data Augmentation method and Feature Similarity based regularized training method achieves the best classification accuracy on standard adversarial attacks such as FGSM and BIM across different UCR time series dataset.  
\end{enumerate}

For models to be useful in the real world, they need to be both accurate on a held-out set of time series data, which we refer to as \textit{clean accuracy}, and robust on corrupted time series, which we refer to as \textit{robustness}.  It is believed that there exists a fundamental trade-off between the two (\cite{TsiprasTradeoff2018}). Our observation has been that though the trade-off exists, it is possible to build robust systems with very minimal or no drop in clean accuracy.

\section{Gradient Based Augmentation Techniques} \label{genadvsampmain}
In the time series classification paradigm, neural networks are maximum likelihood estimators.  The neural network tends to learn the input features that are important for classification, even if those features look incomprehensible to humans.  It has been shown that simple gradient based perturbations in the input signal can cause a trained network to misclassify (\cite{GoodfellowFGSM2014,KurakinBIM2016}).  We first assume a threat model (black-box or white box) and generate adversarial samples based on that model.  We augment original train data with adversarial samples and then train a reference neural network with this data.  We evaluate the trained network on standard adversarial attacks on time series data (\cite{FawazTSCAdv2019}).

\begin{figure}[t] \centering
	\centerline{\includegraphics[width=1.0\linewidth,height=4cm]{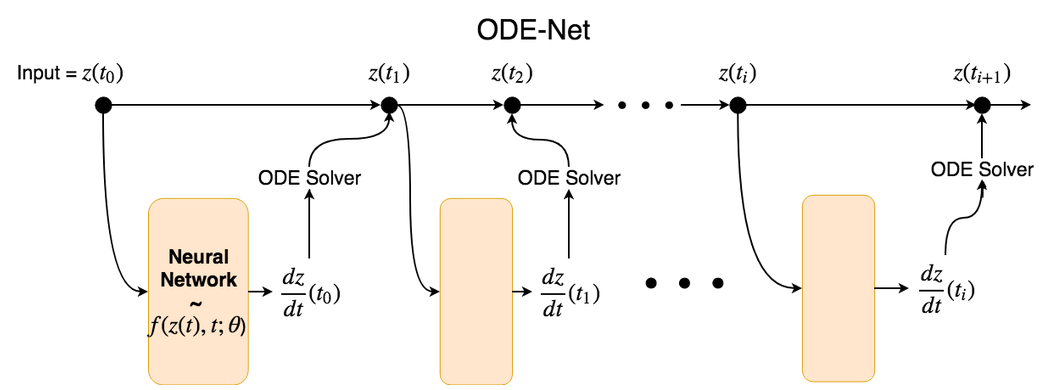}}
	\caption{ODENet building blocks. Parameters of ODENet are shared across times.}
	\label{fig:odenet}
\end{figure}

\subsection{Input gradient based Data Augmentation  and Training method} \label{genadvsamp}
\textbf{STEP-1:} Neural networks were used to model Ordinary Differential Equations(ODE) (\cite{ChenOdenet2018}). We define a time-series $z(t)$ as a sequence of 1-D values $z(t_{0}),z(t_{1}),...,z(t{n})$. We model the continuous dynamics of the input time series using an ODE specified by a neural network(f), parameterized by $\theta$:
\begin{equation}
\frac{dz(t)}{dt} = f(z(t),t,\theta)
\label{eq:1}
\end{equation}
We refer such neural network as \textbf{ODENet} as shown in fig.\ref{fig:odenet}. We use ODE solver as a sequential generative model to predict successive samples of the time-series. More specifically, we compute $z(t_{i})$ from known $z(t_{i-1})$ as:
\begin{equation}
z(t_{i}) = z(t_{i-1}) + \int^{t_{i}}_{t_{i-1}} f(z(t),t,\theta) dt
\label{eq:2}
\end{equation}
We predict the values of $z(t_{1}),z(t_{2}),...,z(t{n})$ by iteratively following \ref{eq:2}. The value predicted at time step $t_{i}$ is treated as the initial value for the computation of $z(t_{i+1})$. We minimize MSE loss between predicted and true sequence, in order to find the optimum parameter set of ODENet,$\theta_{opt}$, which can capture the continuous dynamics of $z(t)$ at any given point as follows:
\begin{equation}
\frac{dz(t_{i})}{dt} = f(z(t_{i}),t_{i},\theta_{opt})
\label{eq:3}
\end{equation}

Adam optimizer with a learning rate of 0.0003 and weight decay factor of $10^{-3}$ was used for training the ODENet.

\textbf{STEP-2:} Once the ODENet is trained, we use it to compute the gradient of a timeseries at any given time point using eq. \ref{eq:3}. we add carefully crafted perturbation guided by the direction of gradient at each time point, which allows to preserve the overall structure of the original input time series.

\begin{figure}[t] \centering
	\centerline{\includegraphics[width=1.0\linewidth,height=4cm]{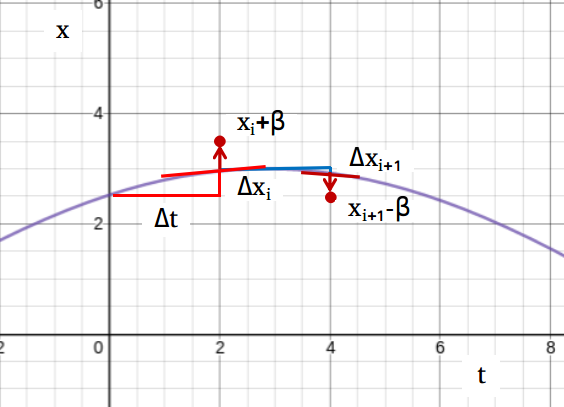}}
	\caption{{\it Input gradient based adversarial sample generation}.  Each point in the timeseries is perturbed by a random number in range [0,$\epsilon$].  The direction of change is determined by the local gradient of $x$.
	}\label{fig:method1explain}
\end{figure}

\begin{figure*}[t] \centering
	\centerline{\includegraphics[width=1.0\linewidth,height=6cm]{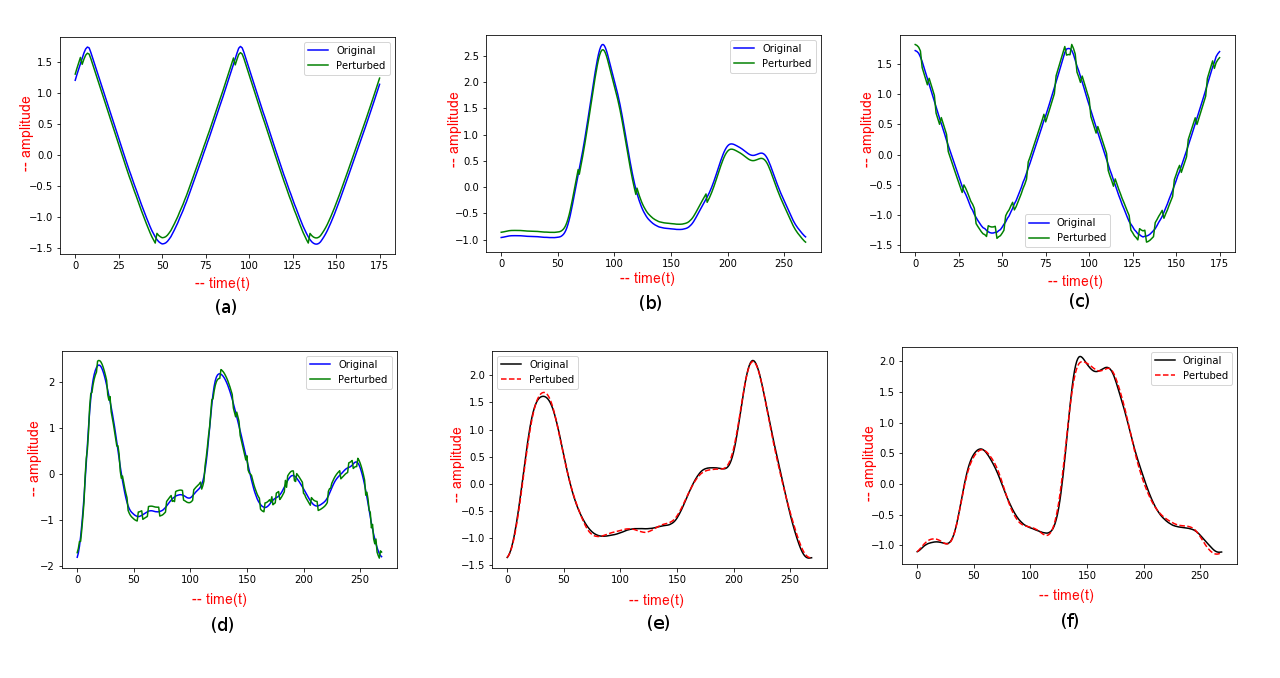}}
	\caption{Two randomly selected samples and their corresponding adversarial samples obtained using \textbf{(a),(b)} {\it Input gradient}, \textbf{(c),(d)} {\it Output gradient}, \textbf{(e),(f)} {\it Spectral Density} based adversarial sample generation techniques.  \textbf{(a),(c) and (e)} are from Adiac time series dataset while \textbf{(b),(d) and (f)} from 50words dataset.}
	\label{fig:methodsex}
\end{figure*}

Accordingly the augmented samples are generated using the below equation:
\begin{equation} \label{dxdtperturb}
z_{aug}(t_{i}) = z(t_{i}) + clamp(\frac{dz(t_{i})}{dt})
\end{equation}

\begin{equation} \label{clamp}
clamp(z) = 
\begin{cases}
z & |z| <\beta \\
\beta & z > \beta \\
-\beta & z < -\beta
\end{cases}
\end{equation}
$z_{aug}(t_{i})$ and $z(t_{i})$ are respectively augmented timeseries and original timeseries at timestep $t_{i}$. $\beta$ is a small positive constant. Note that by perturbing the samples as given in equation (\ref{dxdtperturb}), the change in magnitude of each sample in the augmented timeseries has an upper limit of $\beta$ compared to the original timeseries.  This inherently sets an upper limit on the change in gradient to $2*\beta / \delta t$ as shown in fig  \ref{fig:method1explain}. Figure \ref{fig:methodsex}(a),(b) shows example timeseries data augmented using this technique. We term this augmentation method as \textbf{In-Clamp-Grad}.

\textbf{Training Objective of Classification Network:} We define a multiclass classification setup where the input-label pairs $(z,y) \in (\chi$ x $\lambda$) are sampled from data distribution $D$. A neural network is trained as a classifier whose goal is to predict the class label $y$ for a given input $z$. The training objective of the classifier is to find parameters $\phi$ that minimize the multiclass cross entropy loss:

\begin{equation} \label{eqn:trainmaxcorr}
\min_{\phi} \; \rm I\!E_{(z_{aug},y)}[CELoss(g_{\phi}(z_{aug}, y))]
\end{equation}
Here, $g_{\phi}$ represents the neural network classifier with parameters $\phi$. We augment the input dataset with the perturbed examples obtained using equation (\ref{dxdtperturb}). The labels for the perturbed examples are set to be same as the original example from which it was obtained. Thus, training the neural network classifier with this augmented dataset follows \ref{eqn:trainmaxcorr}. We used Adam optimizer with a learning rate of 0.0002 with a weight decay of $10^{-3}$ for all the experiments involving training the classifier network in this paper. We denote $(z_{aug},y)$ as input-label pairs of augmented dataset.

We also experimented with a slight modification of equation (\ref{dxdtperturb}) as given below:
\begin{equation} \label{dxdtsign}
z_{aug}(t_{i}) = z(t_{i}) + \epsilon* sign(\frac{dz(t_{i})}{dt})
\end{equation}
Here, $\epsilon$ is a random value between 0.0 and 0.33. We term this augmentation method as \textbf{In-Sign-Grad}.

\subsection{Output gradient based Data Augmentation Method}\label{genadvsamp2}
In this case, we assume the adversary has knowledge of the model and its parameters. We perturb the input samples based on the derivatives of classifier output w.r.t input. If $g()$ is the transformation applied by the neural network and $z$ is the input timeseries data, then the derivative of classifier output w.r.t the time variable is governed by the following chain rule:
\begin{equation} \label{eqn:outderivative}
\frac{dy}{dt} = \frac{dg(z)}{dz}\frac{dz}{dt}
\end{equation}

As done in section \ref{genadvsamp}, $\frac{dz}{dt}$ is computed using an ODENet. $\frac{dg(z)}{dz}$	is computed using the automatic differentiation in Pytorch. Using the derivative of classifier output w.r.t the time variable, we define a perturbed time series as below:

\begin{equation} \label{dydt_dxdt}
z_{aug}(t) = z(t) + sign(\frac{dg(z)}{dz})\; abs(clamp(\frac{dz}{dt}))
\end{equation}

where, $sign(\frac{dg(z)}{dz})$ is an indication of the direction of maximum change in classifier output w.r.t input. For cross entropy loss, changing the input along this direction maximizes the loss term. $\frac{dz}{dt}$ is the gradient of input w.r.t time. The function $clamp$ is defined in equation ($\ref{clamp}$). Thus, the direction of perturbation is defined by $(\frac{dg(z)}{dz})$, while the magnitude of perturbation is defined by $\frac{dz}{dt}$ and it is limited by a small positive number ($\beta$), which helps in preserving the overall structure of original time series.
As done in section \ref{genadvsamp}, we augmented the input dataset with perturbed samples and trained a neural network for classification. The trained neural network achieves state-of-the-art results on different UCR time series test datasets with FGSM and BIM perturbations. Figure \ref{fig:methodsex}(c),(d) shows example timeseries data augmented using this technique. We term this augmentation method as \textbf{Out-Sign-Grad}.

Note that in figure \ref{fig:methodsex}, sample generated using output gradient based technique has much more fluctuation compared to sample generated using input gradient based technique. This is intuitive, because in Out-Sign-Grad technique, at each time point, perturbation direction is guided by gradient of classifier output w.r.t input at that point. Hence, such perturbation can behave in arbitrary manner and have no relation with original time series. But in In-Sign-Grad technique, the perturbation direction is guided by the “learned” gradient of the original input time series at that time point. As a result, input gradient based augmented time series preserves the shape of original input time series better compared to output gradient based augmented time series.

To validate the significance of $sign(\frac{dg(z)}{dz})$ in the generation of perturbed samples, we replaced this with random sign for each time step.  

\begin{equation} \label{rand_dxdt}
z_{aug}(t) = z(t) + random([-1,+1])\; abs(clamp(\frac{dz}{dt}))
\end{equation}

A neural network trained with such augmented data, which we termed as \textbf{Rand-Grad}, was found to provide lesser accuracy compared to the network trained using the augmented data obtained using \textbf{Out-Sign-Grad}.

\subsection{Spectral density based Data Augmentation method} \label{sec:specdens}
In the techniques described in previous sections, we introduced perturbations at each timestep of the input timeseries. The perturbations added were functions of derivatives of inputs w.r.t the time and/or derivatives of classifier output w.r.t the input. In this section, we describe the perturbations added based on the frequency domain characteristics of the input signals. The energy of a signal is given by Parseval's theorem:
\begin{equation}
E(z) = \sum_{n=0}^{N-1}|z[n]|^2 = \frac{1}{N}\sum_{k=0}^{N-1}|Z[k]|^2
\end{equation}
Where, $Z[k]$ is the Discrete Fourier Transform of $z[n]$, both of length $N$.

We now obtain a new perturbed Discrete Fourier Transform $Z_{aug}[k]$ as described below.  The sequence $Z[k]$ is sorted in descending order to obtain $Z_{sorted}$.
\begin{equation}
Z_{sorted} = sort(|Z[k]|) 
\end{equation}
We then find the index $C$ such that:
\begin{equation} \label{defnC}
\sum_{k=0}^{C}|Z_{sorted}[k]|^2 \geq 0.9*E(z) > \sum_{k=0}^{C-1}|Z_{sorted}[k]|^2
\end{equation}
We apply perturbations to the frequency components in the sorted sequence ($Z_{sorted}[k])$ which lie above the index $C$. Let $Z_1$, $Z_2$ and $Z_3$ be three consecutive frequency components in $Z_{sorted}[k]$ which lie above the index $C$, while $Z_{1p}$, $Z_{2p}$ and $Z_{3p}$ be the corresponding perturbed components. Then, the perturbed frequency domain components are obtained as:
\begin{eqnarray} \label{freqperturb}
Z_{1p} = Z_1 + \frac{Z_2}{4}, \; Z_{2p} = Z_2 + \frac{Z_2}{2}, \;
Z_{3p} = Z_3 + \frac{Z_2}{4}
\end{eqnarray}
We apply such perturbation on upto 75\% of the components which lie above the index $C$. All components below index $C$ remain same. The perturbed time domain timeseries is obtained by taking inverse DFT of the perturbed frequency domain components. Thus, we only redistribute energy in the frequency components forming less than 10\% of the energy of the signal. Figure \ref{fig:methodsex}(e),(f) shows example timeseries data augmented using this technique. We term such augmentation method as \textbf{Spec-Den}.

\section{Experiments and Results}

\subsection{Network Architecture}

\subsubsection{ODENet} \label{odenetarch}
The timeseries dynamics of the input are modelled using differential equations. A neural network is trained to compute the gradients of input timeseries at any given timestep (eq. \ref{eq:3}).
The neural network $f()$ consists of two hidden layers of 25 neurons. Each hidden layer is followed by batch normalization and ELU activation. The value of the input timeseries at a given time $(z(t_i))$ and the timestep $(t_i)$ form the input to the neural network. The network predicts the gradient of input time series at timestep $(t_i)$.


\subsubsection{Classification Network} \label{classnet}
\begin{figure}[h!] \centering
	\centerline{\includegraphics[width=1.0\linewidth]{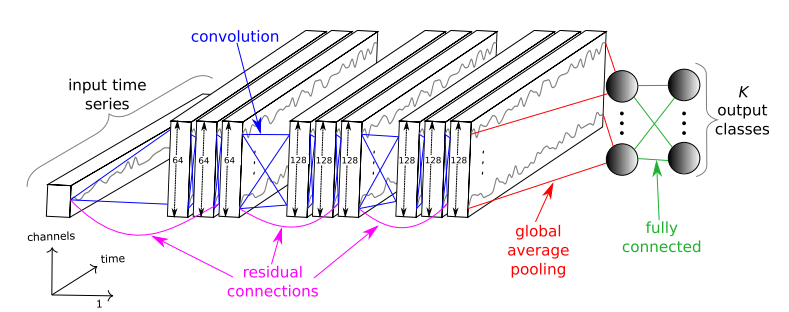}}
	\caption{Reference ResNet architecture of the neural network used for all experiments in this paper (from \cite{FawazTSCAdv2019})
	}\label{fig:resnetarch}
\end{figure}

The neural network for timeseries classification follows the ResNet architecture as defined in \cite{FawazTSCAdv2019}, which is depicted in fig \ref{fig:resnetarch} . The input to this network is a time series of length T. The output of the network is a probability distribution over the K possible classes in the dataset. The network consists of 9 convolutional layers, grouped into three residual blocks. Each layer is followed by a Rectified Linear (ReLU) activation and batch normalization.  This is followed by a global average pooling layer and a softmax classification layer. We retain the same network architecture and train it with the dataset augmented with our proposed methods. This allows us to compare the effectiveness of our techniques with the baseline in \cite{FawazTSCAdv2019}.

\subsection{Datasets}
We use the UCR timeseries dataset (\cite{ChenUCR2015}) for all the experiments described in this paper. Table \ref{tab:result1} summarizes the datasets used, along with the baseline true classification accuracy of the ResNet classifier, baseline adversarial accuracy measured on the test data perturbed using FGSM (\cite{GoodfellowFGSM2014}) and BIM (\cite{KurakinBIM2016}) attack on the ResNet classifier (as reported in \cite{FawazTSCAdv2019}).

Using the techniques described in section \ref{genadvsampmain}, we augment the datasets listed in table \ref{tab:result1}. We use the augmented dataset to train the baseline classification network defined in section \ref{classnet}. The timeseries signals of all the datasets used in our experiments displayed an amplitude range close to 6 units. The value of $\beta$ (equation \ref{clamp}) was chosen to be 0.33 and $\epsilon$ (equation \ref{dxdtsign}) was a random number in range [0.0,0.33] for all the experiments described in this paper. Note that, We choose the value of $\beta$ and $\epsilon$ in such a way, so as to limit the perturbation range to be almost $\frac{1}{10}$th of the amplitude range of the original time series.

\begin{table*}[!hbt]
    \caption[short text]{Comparison of True, Adversarial classification accuracy on networks where training data is augmented using techniques described in section \ref{genadvsampmain} and baseline True, Adversarial accuracy}\label{tab:result1}	
	\centering
	\resizebox{\textwidth}{!}{
\begin{tabular}{|l|l|l|l|l|l|l|l|l|l|l|l|l|}
	\hline
	& \multicolumn{3}{c|}{Baseline} & \multicolumn{3}{c|}{Input Gradient Method} &
	\multicolumn{3}{c|}{Output Gradient Method} &
	\multicolumn{3}{c|}{Spectral Density} \\ \cline{2-13} 
	& \multicolumn{3}{c|}{No Augment} & \multicolumn{3}{c|}{\textbf{In-Clamp/In-sign-Grad}} & \multicolumn{3}{c|}{\textbf{Out-Sign-Grad}} & \multicolumn{3}{c|}{\textbf{Spec-Den}} \\ \cline{2-13} 
	\multirow{-3}{*}{Dataset} & True & FGSM & BIM & True & FGSM & BIM & True & FGSM & BIM & True & FGSM & BIM \\ \hline
	CricketX & 79.00 & 35.4 & 20.8 &
	{\color[HTML]{303498} \textbf{79.49}} & {\color[HTML]{303498} \textbf{47.95}} & {\color[HTML]{303498} \textbf{38.72}} & {\color[HTML]{009901} \textbf{79.49}} & {\color[HTML]{009901} \textbf{72.31}} & {\color[HTML]{009901} \textbf{72.05}} & 79.23 & 46.67 & 37.44 \\ \hline
	CricketZ & 81.5 & 27.7 & 16.2 & {\color[HTML]{303498} \textbf{82.31}} & {\color[HTML]{303498} \textbf{43.33}} &
	{\color[HTML]{303498} \textbf{34.87}} & {\color[HTML]{009901} \textbf{81.79}} & {\color[HTML]{009901} \textbf{42.82}} & {\color[HTML]{009901} \textbf{37.18}} & 81.03 & 40.26 & 33.08 \\ \hline
	50Words & 73.2 & 17.1 & 8.8 & {\color[HTML]{303498} \textbf{74.73}} & {\color[HTML]{303498} \textbf{38.46}} & {\color[HTML]{303498} \textbf{29.23}} & {\color[HTML]{009901} \textbf{76.04}} & {\color[HTML]{009901} \textbf{67.25}} & {\color[HTML]{009901} \textbf{52.97}} & 74.07 & 25.5 & 12.09 \\ \hline
	UWaveGestureLibrary\_X & 78.0 & 32.1 & 11.1 & {\color[HTML]{303498} \textbf{78.28}} & {\color[HTML]{303498} \textbf{46.76}} & {\color[HTML]{303498} \textbf{20.91}} &
	{\color[HTML]{009901} \textbf{79.23}} & {\color[HTML]{009901} \textbf{71.19}} & {\color[HTML]{009901} \textbf{55.05}} & 77.11 & 40.90 & 19.88 \\ \hline
	InsectWingbeatSound & 50.6 & {\color[HTML]{000000} 17.7} & {\color[HTML]{000000} 15.7} & {\color[HTML]{303498} \textbf{50.81}} & {\color[HTML]{303498} \textbf{25.81}} & {\color[HTML]{303498} \textbf{21.72}} & {\color[HTML]{009901} \textbf{53.48}} & {\color[HTML]{009901} \textbf{39.14}} & {\color[HTML]{009901} \textbf{38.13}} & 51.16 & 25.20 & 20.05 \\ \hline
	Adiac & 83.1 & 3.1 & {\color[HTML]{000000} 1.5} & {\color[HTML]{303498} \textbf{83.38}} & {\color[HTML]{303498} \textbf{6.39}} & {\color[HTML]{303498} \textbf{3.84}} & {\color[HTML]{009901} \textbf{83.63}} & {\color[HTML]{009901} \textbf{45.27}} & {\color[HTML]{009901} \textbf{25.83}} & 83.38 & 5.63 & 4.09 \\ \hline
	TwoLeadECG & 100 & 5.3 & 0.4 & {\color[HTML]{303498} \textbf{100}} & {\color[HTML]{303498} \textbf{14.05}} & {\color[HTML]{303498} \textbf{7.64}} & {\color[HTML]{009901} \textbf{100}} &
	{\color[HTML]{009901} \textbf{61.98}} & {\color[HTML]{009901} \textbf{31.69}} & 100 & 14.72 & 7.20 \\ \hline
	UWaveGestureLibrary\_Y & 66.7 & 27.7 & 14.9 & {\color[HTML]{303498} \textbf{67.03}} & {\color[HTML]{303498} \textbf{36.24}} & {\color[HTML]{303498} \textbf{17.64}} & {\color[HTML]{009901} \textbf{67.06}} & {\color[HTML]{009901} \textbf{57.96}} & {\color[HTML]{009901} \textbf{47.35}} & 66.14 & 35.87 & 18.01 \\ \hline
\end{tabular}
	}
\end{table*}

\begin{table*}[!t]
\centering
\caption{Adversarial accuracy on networks trained using additional Feat-Sim regularizer.}
\label{tab:datasets}
\begin{tabular}{|l|l|l|l|p{1cm}|p{1cm}|l|l|p{1.2cm}|p{1.2cm}|p{1cm}|} 
\cline{2-7}\cline{9-11}
\multicolumn{1}{l|}{} & \multicolumn{6}{l|}{ (A): Input Gradient Method + \textbf{Feat-Sim}}                                                                                &  & \multicolumn{3}{l|}{(B):Out-Sign-Grad+\textbf{Feat-Sim}}            \\ 
\cline{1-7}\cline{9-11}
Dataset               & \multicolumn{3}{l|}{In-Clamp-Grad+\textbf{FS}} & \multicolumn{3}{l|}{In-Sign-Grad+\textbf{FS}} &  & \multicolumn{3}{l|}{$\>\>\> \>\>$Out-Sign-Grad+\textbf{FS}}  \\ 
\cline{1-7}\cline{9-11}
                      & True & FGSM & BIM                  & True & FGSM & BIM                  &  & True & FGSM & BIM                   \\ 
\cline{1-7}\cline{9-11}
CricketX                   & 79.23   & 47.18   &  37.44                    & 79.23     & \textbf{51.28} & \textbf{42.05}                     &  & 79.49     & \textbf{75.90}     &  \textbf{76.41}                   \\ 
\cline{1-7}\cline{9-11}
CricketZ                   & 81.54     &  40.00    &  31.03                    & 82.31     & \textbf{45.90}      & \textbf{36.15}                    &  &  81.73     & \textbf{66.41}     &  \textbf{66.15}                       \\ 
\cline{1-7}\cline{9-11}
50Words                      &  74.07    & \textbf{40.66}     & \textbf{30.33}                     &  73.20    & 34.73     & 19.34                     &  & 76.04      & \textbf{69.23}     & \textbf{55.82}                      \\ 
\cline{1-7}\cline{9-11}
UWaveGestureLibraryX                      & 78.42     & \textbf{51.90}     & \textbf{28.25}                     & 78.03      &  47.74    &   24.06                  &  & 78.53      & \textbf{72.47}     & \textbf{62.62}                       \\ 
\cline{1-7}\cline{9-11}
InsectWingBeatSound                     & 49.85     &  26.41    &  22.42                    & 49.29     & \textbf{28.23}    & \textbf{23.38}                    &  & 51.87     & \textbf{43.74}     &  \textbf{42.17}                     \\ 
\cline{1-7}\cline{9-11}
Adiac                      & 79.28     & \textbf{7.16}      &    \textbf{5.09}                  &  82.61    & 6.65     & 4.86  
                       &  & 83.38      & \textbf{46.29}      & \textbf{28.64}                      \\ 
\cline{1-7}\cline{9-11}
TwoLeadECG                      & 100     & \textbf{38.81}     &   \textbf{29.06}                   & 100     &  29.24    & 21.60 
                      & & 100 & \textbf{73.40}   & \textbf{30.38}                            \\ 
\cline{1-7}\cline{9-11}
UWaveGestureLibraryY                      & 66.86     &  36.71    &  20.69                    & 66.95     & \textbf{41.54}     & \textbf{25.43}                     &  & 67.07     & \textbf{59.35}     &  \textbf{58.15}                     \\
\cline{1-7}\cline{9-11}
\end{tabular}
\end{table*}

\subsection{Additional Technique} \label{addtech}

In addition to the techniques already described for adversarial sample generation, we experimented with following regularization technique to improve the classification accuracy. We describe this below:

\subsubsection{Feature Similarity}
Distance metric for large margin nearest neighbour classification was introduced in \cite{Weinberger2009LM} and later used for face recognition and clustering in \cite{Schroff2015FaceNet}. Let $z_c$ be a time series sample of a specific class $c$. Let $z_p$ be the augmented sample obtained by perturbing $z_c$ using techniques described in section \ref{genadvsampmain}.  Let $z^i_{c'}$ be a randomly chosen time series sample of any other class $c' \neq c$. The augmented sample obtained by perturbing $z^i_{c'}$ is given by $z^i_{p'}$. $g(z) \in \mathbb{R^d}$ represents the d-dimensional embedding of the timeseries $z$ obtained by the neural network. Then, the loss function that is being minimized has an additional term (along with  cross entropy) given by
\begin{equation}
L_{FS} = ||g(z_c) - g(z_p)||^2_2 + \alpha - ||g(z_c) - g(z^i_{p'})||^2_2
\end{equation}
where $\alpha$ is a margin that is enforced between positive and negative  pairs.  We also ensured that the  embedding  to  lives on the d-dimensional hypersphere, i.e. $|| g(z) ||^2 = 1$. We term this regularized training strategy as \textbf{Feat-Sim(FS)}.

\subsection{Results}
In this section, we describe and summarize the classification accuracy observed on the reference ResNet based classification network (section \ref{classnet}), where the training dataset is augmented using techniques summarized in section \ref{genadvsampmain}. 

{\textbf{Comparison with baseline: Table \ref{tab:result1}}} reports the comparison between baseline accuracy and the best accuracy achieved using our following proposed augmentation methods: Output Gradient based Augmentation, Input Gradient based augmentation and spectral Density based augmentation.We observe that even the Input Gradient based augmentation methods (equations \ref{dxdtperturb}, \ref{dxdtsign}) significantly improve the adversarial classification accuracies (FGSM and BIM).  For every dataset, we highlight the Input Gradient Based technique that provided highest adversarial classification accuracy with {\color[HTML]{303498} \textbf{blue}} and the Output Gradient based technique that provided highest adversarial classification accuracy with {\color[HTML]{009901} \textbf{green}}. In both cases, the true accuracy was either same or better than the Baseline true classfication accuracy of reference network. Reference network trained with augmented data obtained using Output Gradient based methods performed much better than the ones trained with data obtained using Input Gradient based method. This made intuitive sense because output gradient based method use the gradient of output of the network w.r.t input variable while input gradient based techniques use no knowledge of the network.

\textbf{Comparison between different Input and Output Gradient based Augmentation methods: } \textbf{Table \ref{tab:result2} in Appendix}, we provide detailed comparison between different Input and Output Gradient based Augmentation methods. We observe that different input gradient based techniques were found to provide best adversarial classification accuracies for different datasets. This too made intuitive sense because these techniques are based on the gradients of the input time series itself, and hence it was unlikely that any one technique would give best results for all datasets. We also observe that true and adversarial accuracies achieved using Out-Sign-Grad method is much higher compared to Rand-Grad method.This result justify the significance of $sign\frac{dg(z)}{dz}$ term in equation \ref{dydt_dxdt}.

\begin{figure*}[h!] \centering
	\centerline{\includegraphics[width=1.0\linewidth,height = 7cm]{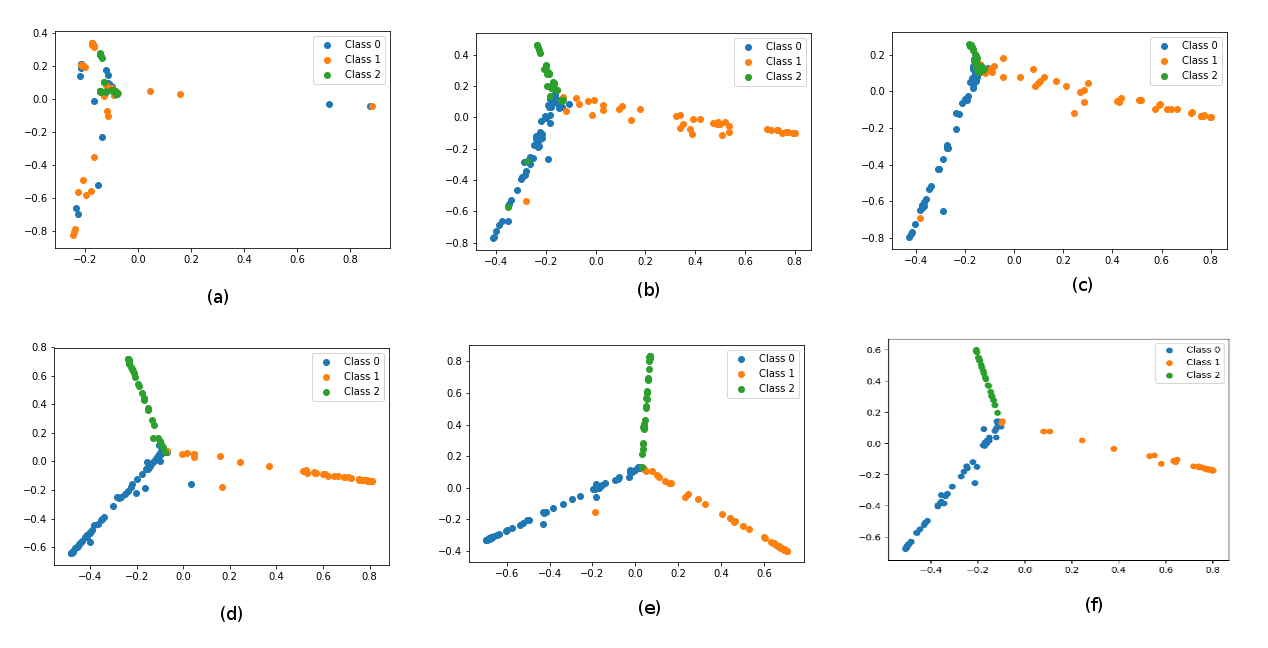}}
	\caption{PCA plots of latent representation of randomly selected samples from the 50Words dataset for network trained with various methods. \textbf{(a)} reference network, \textbf{(b)} In-Clamp-Grad, \textbf{(c)} In-Sign-Grad, \textbf{(d)} Spec-Den \textbf{(e)} Out-Sign-Grad, \textbf{(f)} Out-Sign-Grad + Feat-Sim}
	\label{fig:pca}
\end{figure*}

\begin{figure*}[t] \centering
	\centerline{\includegraphics[width=1\linewidth,height=6cm]{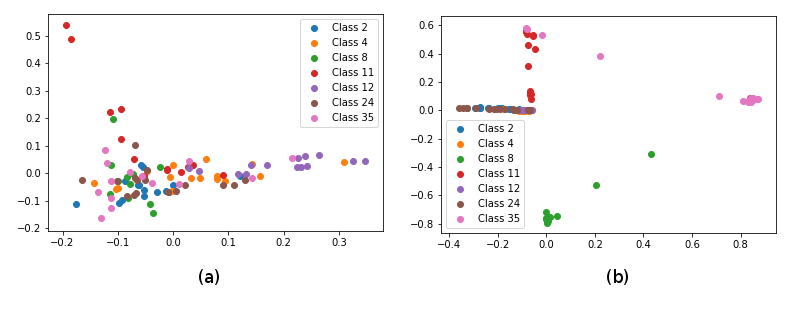}}
	\caption{PCA plots of latent representation of randomly selected samples from the Adiac dataset for \textbf{(a)} reference network and \textbf{(b)}  Out-Sign-Grad + Feat-Sim}\label{fig:pca_conc}
\end{figure*}

\textbf{Adversarial accuracy improvement with additional technique: Table \ref{tab:datasets}:} We also observed that there was significant improvement in the adversarial classification accuracy when we employed additional Feat-Sim regularizer (described in section \ref{addtech}) while training the reference classification network. These are summarized in table \ref{tab:datasets}. It can be seen that a network trained with the combination of  Out-Sign-Grad(Augmentation) and Feat-Sim(Regularizer) (refer table \ref{tab:datasets}) provides the best adversarial classification accuracies along with significant improvements in true accuracies for all cases.

\textbf{Validation}
In this section, we describe an independent technique that we used to validate the augmentation methods described in this paper.

\paragraph{\textbf{Principal Component Analysis}}
In order to visualize the effect of augmentation methods along with additional regularizer technique described in section \ref{addtech} on the latent representation of timeseries samples, we obtained PCA plots of the latent representation of randomly selected samples from the 50Words dataset.  This is shown in figure \ref{fig:pca}.  Here, we can observe that the clustering of samples are better in (b) and (c) (networks trained with augmented samples obtained using input gradient based methods) compared to (a) (reference network).  However, (e) and (f) (networks trained with augmented samples obtained using output gradient based methods) progressively improve the clustering, especially for the samples that lie at the intersection of clusters formed by different classes.  This is inline with the results observed in table \ref{tab:result1} and \ref{tab:datasets}.


\section{Discussion}
 In this paper, we have introduced input gradient and output gradient based augmentation methods. We showed that networks trained using these augmented samples were more robust against standard adversarial attacks without compromising the true accuracy. We have also introduced spectral density based augmentation technique which also improved the robustness of the reference classification network. We have also demonstrated that networks trained with the augmented data are able to cluster the samples in latent space much better than the reference network. However, in (fig \ref{fig:pca_conc}) we have also observed that the samples belonging to certain classes are still inseparable in latent space. We would like to investigate this further and build more robust time series classifiers in our future work. We hope our findings should encourage the researchers to construct more efficient adversarial attack strategy for time series classifiers and should also open the door for new research in the area of adversarial robustness of time series classifier.

\section{Appendix}

\subsection{Detailed results of different augmentation methods}
Table \ref{tab:result2} depicts detailed results of different input and output gradient based augmentation techniques as described in section \ref{genadvsamp} and \ref{genadvsamp2}.
\begin{table}[h!]
    \caption[short text]{Adversarial classification accuracy on network where training data is augmented using techniques described in section \ref{genadvsamp},\ref{genadvsamp2}}\label{tab:result2}		
	\centering
	\resizebox{\textwidth}{!}{
\begin{tabular}{|l|l|l|l|l|l|l|l|l|l|l|l|l|}
	\hline
	& \multicolumn{6}{c|}{Input Gradient Methods} & \multicolumn{6}{c|}{Output Gradient Methods}\\ \cline{2-13} 
	& \multicolumn{3}{c|}{\textbf{In-Clamp-Grad}} & \multicolumn{3}{c|}{\textbf{In-Sign-Grad}} & \multicolumn{3}{c|}{\textbf{Out-Sign-Grad}} & \multicolumn{3}{c|}{\textbf{Rand-Grad*}} \\ \cline{2-13}  
	\multirow{-3}{*}{Dataset} & True & FGSM & BIM & True & FGSM & BIM & True & FGSM & BIM & True & FGSM & BIM   \\ \hline
	CricketX & 79.74 & 44.36 & 31.28 & 79.49 & {\color[HTML]{303498} \textbf{47.95}} & {\color[HTML]{303498} \textbf{38.72}} & 79.49 & {\color[HTML]{009901} \textbf{72.31}} & {\color[HTML]{009901} \textbf{72.05}} & 78.21 & 41.54 & 31.03 \\ \hline
	CricketZ & 82.31 & 40.00 & 30.00 & 82.31 & {\color[HTML]{303498} \textbf{43.33}} & {\color[HTML]{303498} \textbf{34.87}} & 81.79 & {\color[HTML]{009901} \textbf{42.82}} &
	{\color[HTML]{009901} \textbf{37.18}} & 76.92 & 34.87 & 27.95  \\ \hline
	50Words & 74.73 & {\color[HTML]{303498} \textbf{38.46}} & {\color[HTML]{303498} \textbf{29.23}} & 74.29 & 33.19 & 16.70 & 76.04 & {\color[HTML]{009901} \textbf{67.25}} & {\color[HTML]{009901} \textbf{52.97}} & 74.29 & 33.63 & 26.59  \\ \hline
	UWaveGestureLibrary\_X & 78.03 & 46.18 & {\color[HTML]{303498} \textbf{21.11}} & 78.28 & {\color[HTML]{303498} \textbf{46.76}} & 20.91 & 79.23 & {\color[HTML]{009901} \textbf{71.19}} & {\color[HTML]{009901} \textbf{55.05}} & 76.33 & 56.53 & 34.09  \\ \hline
	InsectWingbeatSound & 50.81 & {\color[HTML]{303498} \textbf{25.81}} & {\color[HTML]{303498} \textbf{21.72}} & 49.29 & {\color[HTML]{000000} 27.27} & {\color[HTML]{000000} 23.23} & 53.48 & {\color[HTML]{009901} \textbf{39.14}} & {\color[HTML]{009901} \textbf{38.13}} & 48.44 & 26.06 & 23.79 \\ \hline
	Adiac & 83.38 & {\color[HTML]{303498} \textbf{6.39}} & {\color[HTML]{303498} \textbf{3.84}} & 82.10 & 6.14 & {\color[HTML]{000000} 4.86} & 83.63 & {\color[HTML]{009901} \textbf{45.27}} & {\color[HTML]{009901} \textbf{25.83}} & 81.07 & 17.14 & 11.76 \\ \hline
	TwoLeadECG & 100 & 13.61 & 6.85 & 100 & 14.05 &
	{\color[HTML]{303498} \textbf{7.64}} & 100 & {\color[HTML]{009901} \textbf{61.98}} & {\color[HTML]{009901} \textbf{31.69}} & 99.74 & 17.38 & 6.85 \\ \hline
	UWaveGestureLibrary\_Y & 67.0 & 35.32 & {\color[HTML]{303498} \textbf{18.31}} & 67.03 & {\color[HTML]{303498} \textbf{36.24}} & 17.64 & 67.06 & {\color[HTML]{009901} \textbf{57.96}} & {\color[HTML]{009901} \textbf{47.35}} & 65.01 & 35.21 & 17.61 \\ \hline
\end{tabular}
	}
	* Not a whitebox technique, but used here for comparison
    \end{table}

\subsection{Bound on energy difference between original and augmented sample augmented using spectral density based augmentation method}
if $Z_1,Z_2 \in R$ with $Z_1,Z_2 > 0$ and $|Z_1| > |Z_2|$, then 
\begin{equation}
|Z_1 + \delta|^2 > |Z_2 + \delta|^2
\end{equation}
Thus, if $Z_1,Z_2,Z_3$ are the magnitudes of three consecutive frequency components from $Z_{sorted}$, then the perturbed components are  given by equation \ref{freqperturb}.  The energy difference in these components introduced due to this operation is given by:
\begin{equation}
E_{\delta} = (|Z_{1p}|^2 + |Z_{2p}|^2 + |Z_{3p}|^2) - (|Z_1|^2 + |Z_2|^2 + |Z_3|^2)
\end{equation}

To maximize the difference between $Z1$ and $Z1p$, the component added to $Z_1$ (i.e., $Z_2/4$) should be maximum.  But since $Z_1 \geq Z_2$, $Z_2 = Z_1$ provides the maximum difference between $Z_1$ and $Z_{1p}$.  Along the same lines, we find that $E_{\delta}$ is maximized when $Z_1 = Z_2 = Z_3$.

\begin{equation}
max(|Z_{1p}^2-Z_1^2|) = (Z_1 + \frac{Z_1}{4})^2 - Z_1^2
\end{equation}
\begin{equation}
max(|Z_{2p}^2-Z_2^2|) = (Z_1 - \frac{Z_1}{2})^2 - Z_1^2
\end{equation}
\begin{equation}
max(|Z_{3p}^2-Z_3^2|) = (Z_1 + \frac{Z_1}{4})^2 - Z_1^2
\end{equation}

Or,
\begin{equation}
max(E_{\delta}) = 3.375Z_1^2 - 3Z_1^2
\end{equation}
\begin{equation}
\frac{max(E_{\delta})}{E_{original}} = \frac{3.375Z_1^2 - 3Z_1^2}{3Z_1^2} = 0.125
\end{equation}
Where $E_{original}$ is the energy of the three signal components that are perturbed.
Thus, the maximum difference in energy after the perturbation (equation \ref{freqperturb}) is $12.5\%$ of the original energy present in the three components that are perturbed.

We have assumed (equation \ref{defnC}) that the perturbed components form $10\%$ of the signal energy.  Hence, $max(E_{\delta})$ forms $1.25\%$ of the energy of entire signal.

\end{document}